\pdfoutput=1 
\documentclass[UTF8]{article}

\usepackage{PRIMEarxiv}

\usepackage{natbib}
\usepackage{url}
\usepackage{CJKutf8}
\usepackage{graphicx}
\usepackage{caption}
\usepackage{subcaption}
\usepackage{array}
\newcolumntype{L}[1]{>{\raggedright\arraybackslash}p{#1}}
\usepackage{multirow}
\newcommand{\pcr}[1]{\texttt{#1}}

\newcommand{\umark}{\underline}
\usepackage{csquotes}
\usepackage{amsmath}
\usepackage{setspace}

\usepackage{tabulary}
\usepackage{booktabs}
\usepackage{multirow}

\usepackage{hyperref}       
\usepackage{url}            
\usepackage{booktabs}       
\usepackage{amsfonts}       
\usepackage{nicefrac}       
\usepackage{microtype}      
\usepackage{lipsum}
\usepackage{fancyhdr}       
\usepackage{graphicx}       
\graphicspath{{media/}}     

\title{Resolving Regular Polysemy in Named Entities
}

\author{
  Shu-Kai Hsieh, Yu-Hsiang Tseng, Hsin-Yu Chou, Ching-Wen Yang, Yu-Yun Chang \\
  Graduate Institute of Linguistics \\
  National Taiwan University \\
  Taipei\\
  \texttt{shukaihsieh@ntu.edu.tw} \\
}

\begin{document}
\begin{CJK*}{UTF8}{bsmi}
\maketitle

\begin{abstract}
Word sense disambiguation primarily addresses the lexical ambiguity of common words based on a predefined sense inventory. Conversely, proper names are usually considered to denote an ad-hoc real-world referent. Once the reference is decided, the ambiguity is purportedly resolved. However, proper names also exhibit ambiguities through \emph{appellativization}, i.e., they act like common words and may denote different aspects of their referents. We proposed to address the ambiguities of proper names through the light of regular polysemy, which we formalized as dot objects. This paper introduces a combined word sense disambiguation (WSD) model for disambiguating common words against Chinese Wordnet (CWN) and proper names as dot objects. The model leverages the flexibility of a gloss-based model architecture, which takes advantage of the glosses and example sentences of CWN. We show that the model achieves competitive results on both common and proper nouns, even on a relatively sparse sense dataset. Aside from being a performant WSD tool, the model further facilitates the future development of the lexical resource.
\end{abstract}


\section{Introduction}

Automatic resolution of word sense in running texts has been a long-term pursuit of the NLP community \citep{Agirre2006,Navigli2009}. The resolution task, Word Sense Disambiguation (WSD), is to identify the most appropriate word sense given a sentential context. The WSD task is commonly tackled by two approaches: knowledge-based methods and supervised models \citep{Bevilacqua2021}. Knowledge-based WSD relies on a lexical knowledge base, from which a semantic graph or collocation information is derived, and algorithms are applied to identify the sense that best disambiguates the word. While the knowledge-based method is more flexible as no manually sense-labeled dataset is required, it requires a rich lexical knowledge base in the target language \citep{Scarlini2020ares,Agirre2014}. By contrast, the supervised model is trained on sense-labeled corpora and leverages machine learning algorithms, especially the recent deep learning language models. \citep{Scarlini2020sensembert}

With the advance in technological development in NLP, recent computational semantic studies have witnessed the paradigm shift to the \textit{representation learning}~\citep{liu2020representation}, and the development of WSD is no exception. Representation learning has emerged as an alternative unsupervised approach to feature extraction with the continued success of deep learning. In the paradigm of \textit{representation learning}, features are extracted from unlabeled data by training a neural network on a secondary, supervised task. For example, Google's {\tt word2vec} and its Facebook improvement {\tt FastText} rose to the prominent application of representation learning for symbolic data from the large-scaled textual content in NLP~\citep{Mikolov2013,Bojanowski2017}. 

It is worth noting that in the distributional semantic models\citep{Ferrone2020}, the representation unit is the orthographic (lemma) form, i.e., one word form corresponds to one distributional representation. However, this \textit{type-based} approach `requires explicit composition rules in order to express meaning in context, thus it is intrinsically inapt for the modelling of polysemous words and of polysemysation processes' (quoted from~\citealp{Giulianelli2020}).

Thanks to the latest advances in neural network architectures, context-dependent word features can be embedded into language modeling, e.g., ELMo~\citep{Peters2018}, GPT~\citep{Radford2019}, and BERT~\citep{Devlin2018bert}, among which BERT was shown to achieve the SOTA performance on various downstream tasks. One of its novelties lies in self-supervised training tasks. Specifically, masked language modeling allows the model to `fuse' both left and right contexts; the next sentence prediction task enables the model to capture the relations between sentences. 

In this context, \citep{Levine2019sensebert} further proposes a word sense-informed model called {\tt SenseBERT} to boost the semantic awareness of the existing model. This model is pre-trained to predict the WordNet supersenses, significantly improving lexical understanding, as demonstrated by experimenting on the SemEval WSD task. \citet{Loureiro2019} also constructs word sense embeddings using WordNet's graph to enhance coverage. These findings corroborated our previous exploratory analysis that the sense-aware computational model could be obtained by harnessing external \textit{linguistic knowledge source} with little human annotation or weak supervision. 

It is interesting to note that the WSD task primarily focuses on disambiguating common words, while the disambiguation of \emph{proper nouns} is conventionally configured as another independent module comprised of Named Entity Recognition (NER) and Entity Linking (EL) \citep{Moro2014} (or Named Entity Disambiguation (NED) in general). In terms of NLP task specification, Named Entities are defined as real-world objects that can be denoted by a proper noun which is associated with an entity type such as Person, Product, and Place. A \textit{mention} is a span of text that refers to a named entity in a given text. A mention is often ambiguous because it can refer to different entities, and thus the whole processing involves three subprocesses: identification of these units, categorization of their entity types according to predefined taxonomy, and linking disambiguation which allows to resolve the reference~\citep{nouvel2016named}.

On the surface, WSD and NED are similar but can be distinguished from at least two aspects. First, WSD and NED have different \emph{sense inventories}. WSD often requires a dictionary or WordNet-like lexical resources, and NED relies on an encyclopedia or other knowledge resource to which the proper nouns are resolved. Secondly, proper nouns to be disambiguated are not necessarily complete \emph{mentions}; by contrast, a to-be-disambiguated word is always a fully specified word form \citep{Chang2016}. However, despite these differences, the two tasks are very similar; i.e., they disambiguate words against a reference inventory given their surrounding contexts.

Linguistically, the distinction between common and proper nouns is not always clear-cut. A proper noun is assigned to an \emph{ad hoc} referent in an \emph{ad hoc} name-giving act, resulting in a `proper name' \citep{Langendonck2008}. The proper name does not have a lexical meaning or conceptual meaning, i.e., `the qualities that the referent must possess to be able to be designated by the word in question' \citep{Peterson1989}. The definition seemingly assigns a monoreferential status to proper names, which leaves no ambiguities upon entity linking. However, proper names could be \emph{appellativized} and used as a common word. The appellativization could be done by introducing the indefinite article and adding restrictive modifiers. For example, \emph{I would like to possess a Van Gogh.} The \emph{Van Gogh} is a proper name but is used to indicate the painting by Van Gogh.

Appellativation reveals a crucial aspect of lexical ambiguity. The \emph{Van Gogh} in the example still pertains to the 19-century Dutch artist, to which a mention would connect by a competitive entity linking algorithm. However, in this case, the lexical ambiguity persists as there are different aspects or facets \citep{Langacker1984, Falkum2015} of the referent. That is, the mention could denote either the artist or the paintings he painted. Linguistically, the ambiguities can be observed across different proper names, e.g., Mozart, Shakespeare, and Monet. The sense alternations are \textit{regular}, that is, they are instances of \emph{regular polysemy} \citep{Apresjan1971,Asher2011}.

This paper addresses the lexical ambiguity resolution of common and proper nouns in Mandarin Chinese. For the common nouns, we use Chinese Wordnet (CWN) \citep{Huang2010} as the sense inventory. The CWN is a Chinese lexical semantic resource following the principles of English WordNet (i.e., the Princeton WordNet) but independently compiles the sense inventory and creates sense relations by lexicographers from scratch. CWN's database currently have 11,773 lemmas, 29,384 senses, 94,514 example sentences, and 59,993 semantic relations. Common nouns will be disambiguated based on the CWN sense inventory. On the other hand, the ambiguities of proper nouns are modeled as regular polysemy. We first collect the example sentences from a social media corpus and categorize the polysemous phenomena into seven different types of regular polysemy (dotted types in  \citep{Pustejovsky1995}. We then manually sense-labeled the common and proper nouns against their respective sense inventory and trained a unified supervised model for both types of nouns.
 

The remainder of this paper is organized as follows. Section 2 gives a review of related works on regular polysemy. Section 3 introduces our dataset for word sense disambiguation and regular polysemy of proper nouns. In Section 4, we build and experiment on a unified model which disambiguates common and proper nouns simultaneously; the experiment results and observations are also discussed. Finally, Section 5 describes future works and concludes the paper. 

\section{Related Works}
\label{sec:related-works}

Modeling lexical semantics has been central to much recent work on computational semantics. One line of such research is centered on the treatment of polysemous behavior. The concept of polysemy should first be distinguished from that of monosemy and homonymy. While monosemy simply describes a lexical word having one clear denotation, homonymy displays an external coincidence where two or more unrelated meanings correspond to an identical form. A polysemous word, on the other hand, requires each meaning to be related to at least one other meaning \citep{Apresjan1971,Falkum2015}. While the WSD task primarily focused on the sense enumeration model of polysemy, regular polysemy provides another perspective on sense alternation \citep{Boleda2012}. We first review the literature on regular polysemy and how they are modeled in computational linguistics.

\subsection{Regular polysemy in sense alternations}

Polysemy is regular if there are systematic sense alternations shared across a set of polysemous words. Specifically, \citet{Apresjan1971} defines regular polysemy as follows:

\begin{displayquote}
\indent Polysemy of the word A with the meanings $a_i$ and $a_j$  is called regular if, in the given language, there exists at least one another word B with the meanings $b_i$ and $b_j$ , which are semantically distinguished from each other in exactly the same way as $a_i$ and $a_j$ and if $a_i$ and $b_i$, $a_j$ and $b_j$ are nonsynonymous.
\end{displayquote}

\noindent The systematicity in polysemous words is also studied under different names, such as logical polysemy \citep{Asher2011, Bekki2013}, complementary polysemy, logical metonymy \citep{Pustejovsky1995}, or inherent polysemy \citep{Vicente2021}, each with distinct but closely-related theoretic considerations. For example, \citet{Pustejovsky1995} addresses complementary polysemy in broader contexts, arguing that it refers to two types of sense complementarity: one changes lexical category as its meaning alters, while the other preserves the lexical category. The former has examples such as {\it farm} (with a verbal sense and a nominal sense), while the later is termed {\it logical polysemy}, defined as `a complementary ambiguity where there is no change in lexical category, and the multiple senses of the word have overlapping, dependent or shared meaning' (p. 28). Some classic examples of logical polysemy in English include \texttt{animal.meat} (e.g., \textit{The \underline{chicken} runs fast}; \textit{The \underline{chicken} is delicious.}), \texttt{producer.product} (\textit{\underline{Tesla} has released its financial report}; \textit{I drive a \underline{Tesla}}), and \texttt{physical.information} (\textit{The \underline{book} is heavy}; \textit{The \underline{book} is interesting}). 




Linguistic tests are proposed to test whether the word in question is a logical polysemy. The underlying rationale is that the different senses of a regular polysemous word could be grammatically connected with a conjunction, and could be referred to with an anaphoric pronoun. Consider the two examples below \citep{Falkum2015}: 


\begin{enumerate}
    \item \underline{Lunch} was {\it delicious} but {\it took forever}. \label{ex1}
    \item  The \underline{book} is {\it boring}. Put it on the top shelf. \label{ex2}
\end{enumerate}

Example (\ref{ex1}) passes the linguistic test of conjunction, as the two predicates are connected by the conjunction {\it and} within one sentence, with {\it delicious} pointing to the physical, edible food and {\it took forever} denoting the event of having lunch. On the other hand, (\ref{ex2})  passes the anaphoric-binding test since its first sentence triggers the {\it information} sense of {\it book}, and the second sentence successfully uses a anaphoric pronoun to refer to its {\it physical} sense. \citet{Pustejovsky1995} proposes to call this kind of typing distinction dot-object, where simultaneous predications select two different senses of the polysemous word in one single sentence. That is, the dotted-type of the logical polysemy, \emph{book}, is \texttt{information.physical}.

However, not all dotted-typed polysemous words pass linguistic tests. Some polysemous words follow \citet{Apresjan1971}'s definition but fail the conjunction test, such as \citep{Nunberg1979,Copestake1995}:


\begin{enumerate}
    \item The \underline{ham sandwich} wants a coke. \label{hama}
    \item The \underline{french fries} is getting impatient.\label{hamb}
    \item *The \underline{ham sandwich} wants a coke and has gone stale.\label{hamc}
\end{enumerate}

Both \emph{ham sandwich} and \emph{french fries} are \texttt{food.customer} dot objects, but they cannot pass the linguistic test (\ref{hamc}). \citet{Asher2011} argues that logical polysemy should be distinguished from a broader concept of regular polysemy: the former could pass the co-predication and anaphoric binding tests, while the later may not robustly does so. Logical polysemy and the regular polysemy in general are argued to signify different linguistic considerations and ontological commitments \citep{Arapinis2015,Vicente2021}. Here, however, we follow the general definition of \citet{Apresjan1971} for regular polysemy, and used dot-type~\citep{Pustejovsky1995} as notations to represent different polysemous types.

The dot object captures the regularly polysemous behavior of both common and proper nouns. Indeed, Many of the dot types studied in the literature are particularly pertinent to proper names. Specifically, Beethoven and Shakespeare are instances of \texttt{author.work-of-art} dot object; Honda and New York Times are \texttt{producer.product}, London, Brazil are \texttt{region.people.institution} \citep{Pustejovsky1995}. Namely, the proper names exhibit the polysemous phenomena in language use, even though they are conventionally considered denoting to an \emph{ad hoc} external referent in the real world. The multiple meanings of proper names may be introduced from the name bearer associatively or from the category it belongs to. Particularly, the proper names could be \emph{appellativized} and act like a more complex common noun (appellative). The examples below show how proper names could act like common nouns \citep{Langendonck2008,Vicente2021}.


\begin{enumerate}
    \item \underline{New York} kicked the mayor out of office.
    \item \underline{Brecht} was a communist but is still represented in many theatres in the world.
    \item There we are, him as \underline{L.A.} as a person could get.
\end{enumerate}

Addressing the proper names' polysemous behaviors in light of dot objects is especially pertinent. As regular polysemy could be systematically described with a relatively small set of dot objects, one would not need to enumerate all potential senses of proper names. However, the issue is how to computationally model regular polysemy. Past studies have attempted different approaches, including employing decision trees and ensemble classifiers on linguistic feature sets, word embeddings, sense embeddings, or a complete distributional model for logical polysemy \citep{Boleda2012,DelTredici2015,Lopukhina2016,Chersoni2017}. Each model captures essential aspects of regular polysemy; nevertheless, few of them directly address the lexical ambiguity resolution, especially one of the proper names. On the other hand, models of disambiguating common words, i.e., the WSD task, are less focused on the regular polysemy and dot objects. One possible reason is that the lexical ambiguities of common words are primarily homonymy and irregular polysemy. Consequently, an enumerated approach is preferred, which results in different tasks and models. However, the recent development of contextualized language model shows a possible convergence of computational approaches between disambiguating against a sense enumerated inventory and the regular polysemy described by dot object \citep{Haber2021,Bevilacqua2021}.

\subsection{Word sense disambiguation}

Word sense disambiguation is the NLP task that resolves lexical ambiguities in running texts. In contrast to regular polysemy having systematic sense alternation across lemmas, other polysemy or even homonymy are often \emph{irregular} or \emph{accidental polysemy}, in which `the semantic distinction between $a_i$ and $a_j$  is not exemplified in any other word of the given language' \citep{Apresjan1971,Falkum2015}. That is, there are no systematic relationships among multiple senses that can be described by dot objects. The most straightforward way to represent such sense alterations is to assume that there is a (linguistic or psychological) sense inventory listing all possible senses of a word. Disambiguating the polysemy or homonymy is then identifying the best lexical sense entry given the context from the sense inventory. Although the enumerated approach is not without dispute from the theoretical ground \citep{Sandra1998,Khalilia2021}, WSD models have rapidly evolved and achieved performance comparable to human raters in the past decade \citep{Raganato2017,Maru2022}.

There are at least two common approaches to building a WSD model: a knowledge-based approach and a supervised approach. The knowledge-based approach leverages the existing large-scale knowledge resources to infer the word senses in the context. The lexical resources (e.g., WordNet, ontologies, collocations) not only serve as a sense inventory but additionally provide the relationships among senses from which the algorithm could deduce the word senses \citep{Bevilacqua2021, Navigli2009}. For example, the classic Lesk algorithm \citep{Lesk1986} uses the collocation to compute the \emph{gloss overlap}; specifically, the model calculates the number of shared words for each pair of the sense gloss and the word in context. The sense whose gloss shares the most context words is picked by Lesk algorithm to be the correct sense. There are other algorithms that take advantage of path similarity, which measures the semantic similarity among the senses \citep{Rada1989,Agirre1996,Jiang1997,Leacock1998}. The underlying idea is that the correct assignment of senses have the highest semantic similarities. The same idea also applies to the discourse level. The relationships between the words in the text become a \emph{lexical chain}, in which the semantic similarities are evaluated by their path similarities and the number of relationship \emph{turns} (e.g., from hyponymy to hypernymy) in the chain \citep{Hirst1998,Galley2003}. Furthermore, some algorithm exploits the full sense graph structure comprising all senses of the context words. The centrality measures are computed for each sense node in the graph, and the most relevant nodes are selected as the correct sense assignment \citep{Mihalcea2004,Agirre2014}.

On the other hand, the supervised approach induces a classifier from the sense-annotated dataset. The annotated text is transformed into an input feature vector, from which the classifier learns to assign the correct sense or the class label. Traditionally, the feature vector comes from the preprocessing stages of the NLP pipeline, e.g., tokenization, POS tagging, chunking, and syntactic parsing \citep{Manning2014}. Then, a classifier is built to map these feature vectors into the correct sense labels with machine learning algorithms (e.g., SVM, ensemble method, neural network) \citep{Zhong2010,Navigli2009}. The landscape of WSD models changes significantly after the introduction of word embeddings and contextualized language models. The word embeddings are distributional semantic representations directly learned from the syntagmatic textual context from unlabeled corpus \citep{Mikolov2013,Pennington2014,Collobert2008}. Feature-based WSD model was found to have significant performance improvement when augmented with word embedding features \citep{Iacobacci2016,Raganato2017}. Moreover, the contextualized embeddings extracted from BERT or ELMo \citep{Devlin2018bert,Peters2018} show even more significant improvement over the word embeddings counterparts. Importantly, even without explicitly updating model parameters for WSD task, the correct sense assignment can be identified through the \emph{1-nn vector-based methods} \citep{Peters2018}. That is, the sense having the most similar contextualized embedding with the target word's embeddings is the predicted sense. Different models differ in how they leverage external resources to create high-quality latent representations, either exploiting semantic relations from lexical resources \citep{Loureiro2019,Wang2020} or extracting usage examples from encyclopedia resources \citep{Scarlini2020sensembert,Scarlini2020ares}. 

Another supervised approach to WSD is to exploit gloss information \citep{Blevins2020}. GlossBERT \citep{Huang2019} formulates the WSD problem into a sequence classification task, in which each target word's sense corresponds to one sequence, which is composed of the target's \textit{context} and the sense's \textit{gloss}. The \textit{context} is further augmented by a \emph{weak supervision} signal, which in implementation is a pair of quotation mark around the target word. The model is then fine-tuned to map the contextualized embeddings of the sentence token (\texttt{[CLS]} token) into probabilities of binary responses (\emph{correct} or \emph{incorrect} sequence). In testing, the sequence with the highest \emph{correct} probability is predicted to be the sense of the target word. While GlossBERT attains competitive results, the sequence classification approach is less computationally efficient than the 1-nn vector-based method, since each sense generates a sequence the model has to decode. Another model, ESCHER \citep{Barba2021a}, further optimizes the gloss-based idea and reformulates WSD into a span extraction problem. In ESCHER, all candidate senses' glosses are concatenated into a sequence, and the model is trained to find the gloss span best matches the target word's sense. That means the model only needs to decode one sequence for each target word, though at the cost of a longer sequence length.

The current state-of-the-art WSD models have achieved over 80\% F1 scores on the standard test datasets, which is close to human performance, as indicated by the inter-annotator agreements among raters. Although issues are raised on what the F1 scores reflect in WSD tasks and how the models might tend to bias toward the senses in training set \citep{Palmer2006,Maru2022}, the success of contextualized embeddings and gloss-based approach shed light on the nature of the WSD task. Specifically, the underlying linguistic phenomena of the WSD task is polysemy or homonymy, and these models show us the embeddings from contextualized language models could be used to better identify the word senses in running texts\citep{Loureiro2019,Haber2021}. Importantly, it hints that the WSD model might be not only a downstream application of a given sense inventory but also a tool helping create a better lexical resource. 

\subsection{WSD with Chinese Wordnet}

Chinese Wordnet (CWN, \citealt{Huang2010}) is the Chinese lexical resource for word sense distinctions and semantic relations. Importantly, although CWN tries to follow the principles of PWN \citep{Fellbaum1998}, the sense distinctions are based on the linguistic usages found in corpora, and a panel of lexicographers manually edits sense glosses and example sentences. This \emph{merge} approach of creating a WordNet \citep{Vossen1998}, which starts in one language and then aligns with PWN, could better capture nuances in the language and its underlying ontology. For example, one of the primary usages of 仁 rén, loosely translated to `being able to sympathetically treat others', is a word rooted in Chinese culture. However, there is no direct equivalent synset in PWN. CWN currently has 11,773 lemmas, 29,384 senses, 94,514 example sentences, and 59,993 semantic relations. 

One challenge of CWN is its coverage. Based on a manually-segmented balanced corpus,\footnote{Academia Sinica Balance Corpus from Academia Sinica, Taiwan \citep{Ma2001}} the overall coverage of CWN is 77\%; contrastingly, the coverage of proper nouns is only 46\%. The numbers might not be unexpected as the proper nouns are the open-ended class that updates rapidly, reflecting the ever-changing external world. Nevertheless, including these proper nouns in CWN with the conventional sense enumeration approach will be inefficient. Another way to address the sense alternations of proper nouns is to leverage the fact that most of them are regular polysemy, which we can systematically describe with dot objects. That is, we do not have to enumerate the senses of each proper noun, e.g., Van Gogh or Mozart. Instead, we could list them as \texttt{artist.work\_of\_art} dot objects, and their actual use in the running text could be disambiguated with a WSD model.

Therefore, we task our WSD model with two goals: disambiguate common lexical words against an enumerated sense inventory, i.e., CWN, and resolve proper nouns as dot objects. The developments of WSD models offer us a wide range of model architectures to choose from. However, the architecture in question must satisfy the following criteria: (1) it must be able to disambiguate against a sense inventory or a dot object; (2) as the semantic relations in CWN are still relatively sparse, the model may have to do without the ontological knowledge; (3) there is no large-scale sense-annotated corpus in CWN, the model needs to leverage what the database already includes. These requirements suggest that a pure knowledge-based approach, while linguistically-motivated, might not be applicable for the present task. The 1-nn vector-based approach, while computationally efficient, is not the best one since it is yet to establish the contextualized embeddings of a dot object across different lemmas. By contrast, the gloss-based approach is the most appropriate because it naturally accommodates the enumerated sense and dot object. Furthermore, CWN has already included high quality glosses for each sense. Among the gloss-based approach, we choose GlossBERT \citep{Huang2019} for its flexibility as our model architecture to disambiguate common lexical words and proper nouns.

\section{Annotations and Datasets}

\begin{figure}
\centering
\begin{subfigure}[b]{0.45\textwidth}
    \centering
    \includegraphics[height=0.75\textwidth]{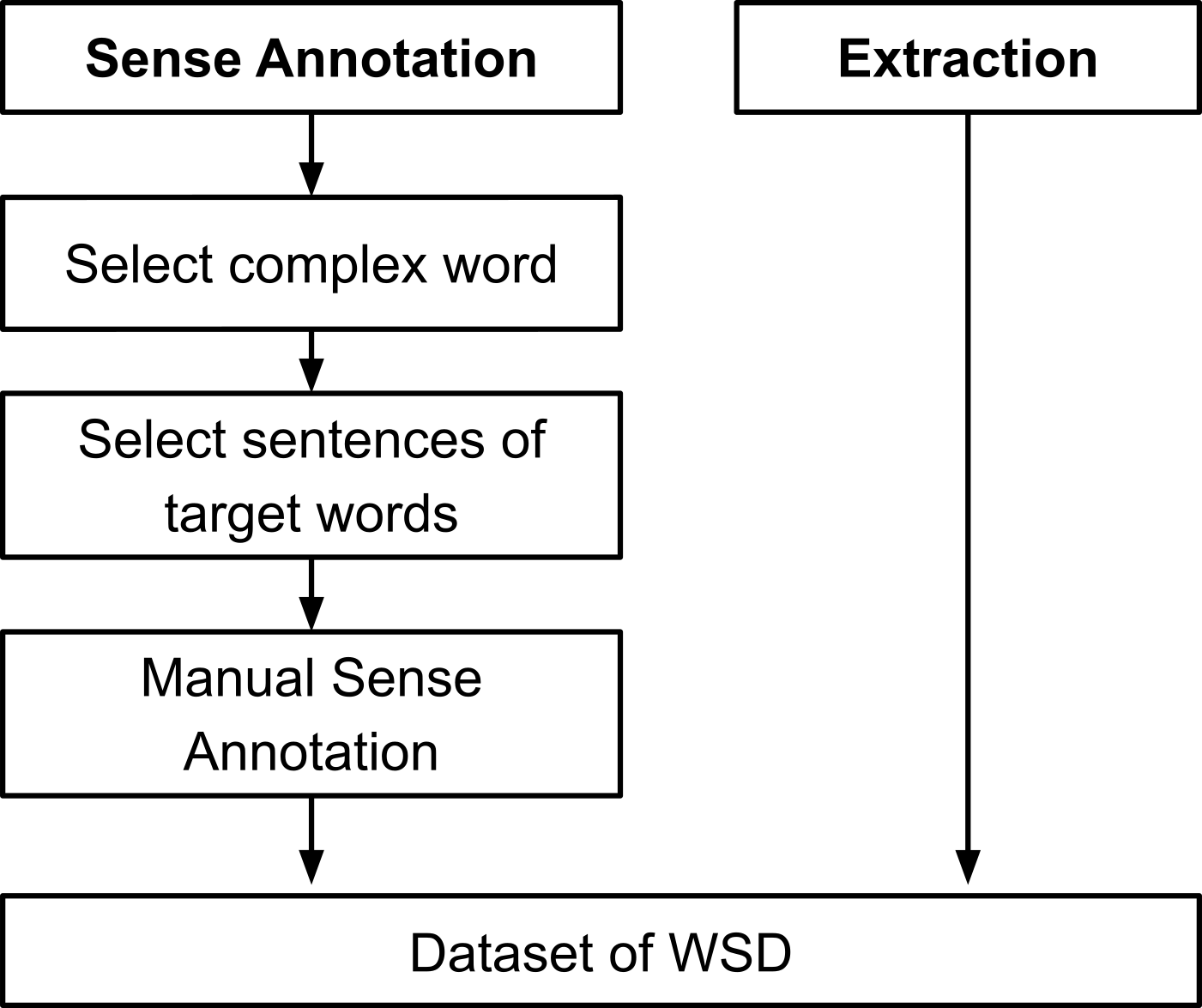}
    \caption{Constructing the dataset of WSD.}
    \label{fig:wsd-proc}
\end{subfigure}
\hspace{0.1cm}
\begin{subfigure}[b]{0.45\textwidth}
    \centering
    \includegraphics[height=0.8\textwidth]{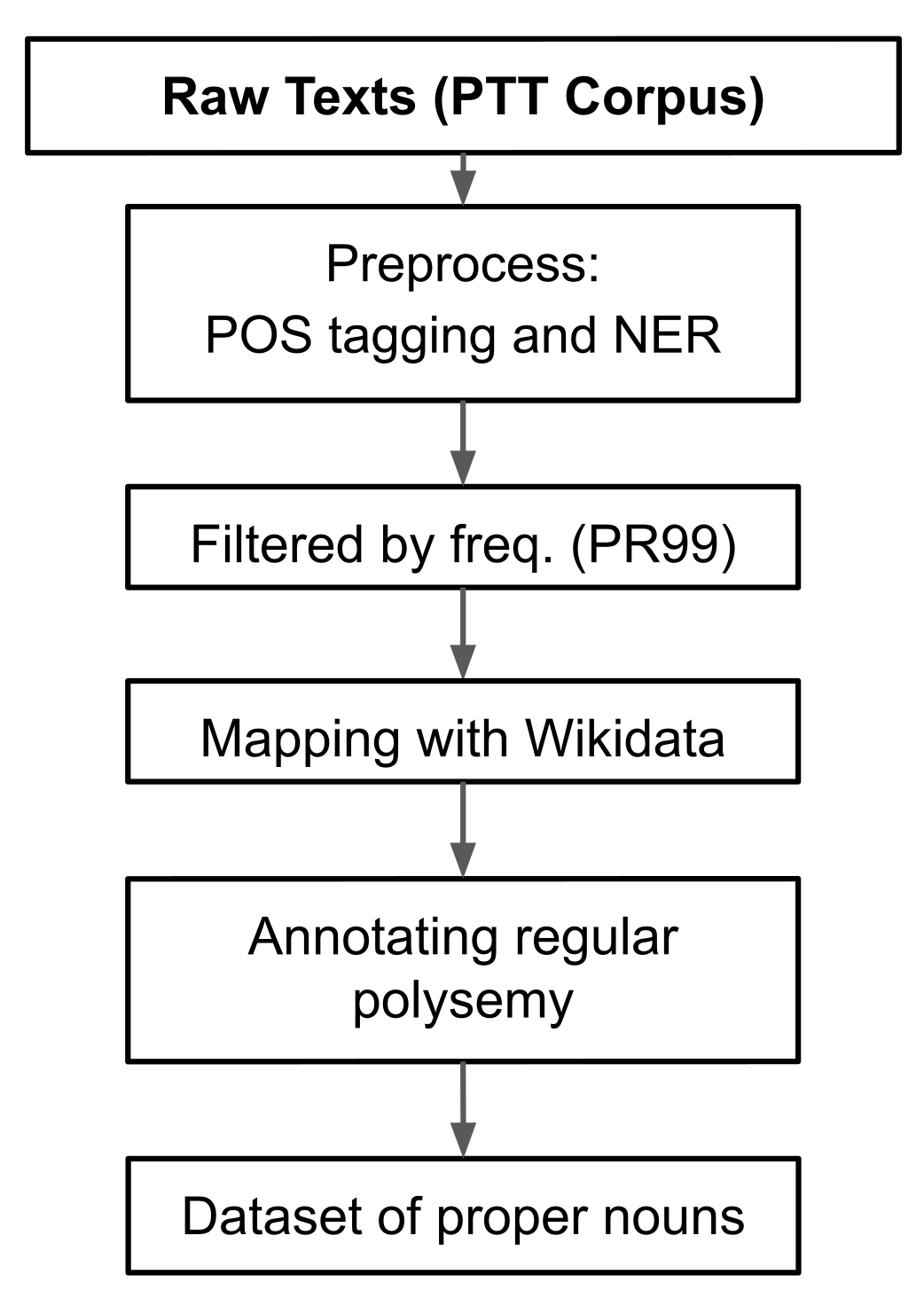}
    \caption{Constructing the dataset of proper nouns.}
    \label{fig:rp-proc}
\end{subfigure}
\caption{The workflows of constructing the datasets.}
\label{fig:dataset-proc}
\end{figure}

To build a WSD model for lexical disambiguation of both common words and proper nouns in Chinese, we compile two datasets: (1) a dataset of WSD and (2) a dataset of proper nouns. The WSD dataset uses the existing CWN sense definitions and example sentences, and includes additional example sentences with manual sense annotations. The WSD dataset follows the conventional sense enumeration approach to describe polysemy. On the other hand, the dataset of proper nouns represents the sense alternations by dot objects. We select proper nouns that are regularly polysemous in the corpus, group them into seven dot objects, and annotate their type classes. The overall procedures of building the datasets are shown in Figure \ref{fig:dataset-proc}. \footnote{The code and datasets are available at the anonymized Github repo: \url{https://anonymous.4open.science/r/proper-wsd-3F6F/}}

\subsection{Dataset of WSD and sense annotation}

Each example in the WSD dataset includes one sentence containing one target word, along with the word's annotated sense and the part-of-speech. Part of these examples is extracted directly from the Chinese WordNet. The sense inventory of CWN contains not only the sense distinctions but the example sentences, which we use as disambiguated contexts for given target words. 

However, the word senses in CWN are carefully differentiated and result in fine-grained sense distinctions. As such, words with complex lexical semantics often contain more than 20 senses. Although CWN provides example sentences for each of them, the limited number of examples may not fully capture the collocation environments, especially for the fine-grained senses. The problem is particularly relevant when considering the usage aspects of these senses: some senses occur more frequently than others; thus, the sentential context may be underrepresented in the CWN database. Therefore, we collect more examples of these complex and fine-grained senses by manually sense annotations.

To this end, we define a word with more than 10 senses in CWN as \textit{difficult word}, and identify 113 difficult words as the target words in annotations. Among them, the word with the fewest senses is 人 rén `human, people, body, etc.' (11 senses), and the word with the most senses is 打 dǎ `hit, dial, dozen, etc.' (125 senses). Sentences containing these target words are extracted from Academia Sinica Balanced Corpus~\citep{Ma2001}. Next, we exclude the sentences with the following criteria: (1) target words are used as proper nouns, domain terms, transliterations, or embedded in multiword expressions or dialectics; (2) segmentation errors occur in the target words. There are 27,012 sentences selected and annotated. The average length of the annotated sentence is 9.77 words.

Six annotators, who are native Mandarin speakers and major in linguistics or TCSL (Teaching Chinese as Second Language), are recruited for sense annotation. Each annotator is assigned a different set of target words. To help annotators familiarize themselves with the sense distinctions, they are asked to categorize the senses into a \emph{sense tree} before labeling senses. There is no specific rule for sense tree building, but annotators are advised to keep the children's size of each node fewer than five. A post hoc analysis of annotators' sense tree shows the average children size of each tree node is 3.47, and the average tree depth is 2.77. Annotators work individually on a web-based annotation interface, which helps them build sense trees, keep track of their work progress, and store the annotation results. Weekly meetings are held to ensure consistent criteria across the annotators.

The annotation dataset contains 28,836 examples. The sense distribution follows the right-skewed pattern commonly observed in corpus data. The most frequently-used senses of each lemma have already accounted for 49\% of the examples. Moreover, among the senses of these 113 target words, only 57\% of the senses are observed in the sampled sentences. That is, some senses might be rarely seen or only used in specific text genres.

Combining the CWN-extracted and human-annotated entries (Figure \ref{fig:wsd-proc}), the dataset of WSD comprises 45,784 examples. While this dataset provides sufficient materials for WSD model of common lexical words, we still require the polysemy data for proper nouns. Therefore, we proceed to construct the proper noun dataset.

\subsection{Dataset of proper nouns and dot objects} \label{ssc: RP Dataset}



The dataset of proper nouns comprises entries denoting an \textit{ad hoc} external referent while having polysemous behaviors. We consider these proper nouns regular polysemy and systematically describe them using a set of dot objects. Therefore, each entry is composed of a target proper noun, the sentential context in which the target occurs, and its type class given the dot object. Therefore, we first define the seven dot objects relevant in our case, select the proper nouns, and manually annotate the type classes in each sentential context.

\subsubsection{Defining dot objects}
\label{sec.dot.objs}
Seven dot objects used in this dataset are listed below. These dot objects are observed in our corpus (see Section \ref{sec.rp.data} for detail), proposed to describe the regular polysemy in both common and proper nouns, and are thoroughly discussed in the literature \citep{Asher2011,Falkum2015,Vicente2021,Pustejovsky2005,Pustejovsky2008}:


\begin{itemize}
    \item information.physical (Info.Phy)
    \item location.organization (Loc.Org)
    \item organization.human (Org.Hum)
    \item organization.information.physical.human (Org.Info.Phy.Hum)
    \item organization.location.human (Org.Loc.Hum)
    \item physical.event.human (Phy.Evt.Hum)
    \item producer.product.location (Prcr.Prct.Loc)
\end{itemize}

Some of these dot objects are explicitly associated with proper names in the literature. For example, \texttt{producer.product} are related to the regular polysemy of company names in the literature, such as \textit{Honda} \citep{Pustejovsky2008}. In contrast, some dot objects are discussed in the context of common nouns, but they have clear relations to a concept class to which a set of proper names belong. For example, \textit{magazine}, \textit{newspaper} are classic examples for regular polysemy, and they are often associated with the dot objects (or complex types) of \texttt{organization.information.physical} \citep{Pustejovsky2005},  \texttt{organization.location.human} \citep{Vicente2021}, \texttt{organization.physical} \citep{Asher2011}. Therefore, we could associate proper nouns denoting specific newspaper instances as \texttt{Org.Info.Phy.Hum} dot objects.

Slight adjustments have been made to the dot object based on our corpus evidence and the characteristics of proper nouns. Take the type \texttt{producer.product.location} for example. A third sense \texttt {location} is added to the dot object \texttt{producer.product}, since the sense referring to the location of a company frequently and regularly appears across different proper names within the semantic class. For example, the proper nouns of producers observed in our corpus are {\it xīng bā kè} (“Starbucks”), {\it kěn dé jī} (“KFC”), or {\it mài dāng láo} (“McDonald's”). They often function as references to the location of the stores, aside from denoting the producer or its products. 

Another observation in our corpus is the close relationship between type class \texttt{human} and \texttt{organization}. \texttt{Human} represents a specific individual, while \texttt{organization} has a related but distinctive sense with a lower degree of individuality. They are frequently observed in collaboration with the dot type class \texttt{organization}. Therefore, \texttt{Org.Info.Phy.Hum} and \texttt{Org.Loc.Hum} both have \texttt{human} type class to reflect the empirical observation. All the adjustments have kept the frequency and regularity of the usage patterns in view and are in consensus among four annotators.

\subsubsection{Collecting proper nouns}
\label{sec.rp.data}
The corpus we observed and compiled the dataset of proper nouns is from PTT Corpus~\footnote{The PTT corpus is a very popular social forum in Taiwan. The topics are organized as different \textit{boards}. We select 20 boards: Horror, Hate, marvel, WomenTalk, Gossiping, JapanMovie, HatePolitics, Tainan, LGBT\_SEX, BabyMother, NTU, nCoV2019, data, sex, Food, Kaohsiung, marriage, gay, Boy-Girl, joke.}. The extracted data are pre-processed before transformed into the dataset for annotation. Figure \ref{fig:evals-wsd} shows the procedures from the raw texts to the annotated dataset.

We totally extract 925K posts (along with the comments in the posts) from 20 forums in 2020. Named entities are first identified in the texts by the CKIP Named Entity Recognition model \footnotemark{}. There are 18 entity types tagged by the \texttt{CkipNerChunker}, and 28M named entity words are extracted out of the 799M tokens from the texts. In light of the seven dot objects we introduced in Section \ref{sec.dot.objs}, five entity types are particularly relevant: LOC ({\it non-GPE locations, mountain ranges, bodies of water}), ORG ({\it companies, agencies, institutions, etc.}), GPE ({\it countries, cities, states}), PRODUCT ({\it vehicles, weapons, foods, etc.}), WORK\_OF\_ART ({\it titles of books, songs, etc.}). The frequency table of the entity types is presented in Table \ref{tab:dist-of-enttype}.

\footnotetext{\texttt{CkipNerChunker}, https://github.com/ckiplab/ckip-transformers}

\begin{table}[ht]
    \centering
    \caption{Distribution of Entity Types. Numbers in the table are the token counts of each entity types instead of the type frequencies.}
    \label{tab:dist-of-enttype}
    \begin{tabular}{lr}
        \toprule
        Entity Type & Number of Instances\\
        \midrule
         GPE & 6,218,407\\
         ORG & 3,280,016\\
         LOC & 418,620\\
         PRODUCT & 274,084\\
         WORK-OF-ART & 141,840\\
        \bottomrule
    \end{tabular}    
\end{table}

We sample a portion of the entities as the candidates of the annotation data. First, the five entity types are separated into their independent dataset. Next, word type frequencies are calculated for each dataset, and frequencies below the 99th percentile are filtered out; that is, we keep the most-frequent 1\% of the word types. There are 370 unique word selected in the final word list.

\subsubsection{Assigning dot objects to proper nouns}
The selected 370 words are potential candidates for our dataset of proper nouns. However, to identify whether they are regular polysemy and the dot objects of these words, we need to associate each word with its external referents. This step is similar to the entity linking or named entity disambiguation in the NLP pipeline. In our case, the goal is to determine the conceptual (or semantic) class of the proper nouns so that we can assign the dot objects to them.

We use Wikidata\footnote{https://www.wikidata.org/wiki/Wikidata:Main\_Page} as our external reference database. We first check whether each word in our word list has a Wikidata entry. If there is an entry for the extracted word, the word is considered a named entity with sufficient usage and prevalence. Words with no entries are removed from further analysis. As a result, there are 38 words removed for no Wikidata entry. These removed words are too-specific terms, such as 國冥黨 \textit{guó míng dǎng}, the pejorative nickname playing with the sound of the second character of 國民黨 \textit{guó mín dǎng} (`Chinese Nationalist Party'), one of the political parties in Taiwan. Other removed terms might be too regional and not ubiquitous enough to have their own Wikidata, such as 高雄地檢署 \textit{gāo xióng dì jiǎn shǔ}, the district prosecutor's office in Kaohsuing, the city in southern Taiwan. On the other hand, there are 15 words with two distinctive entries. All 15 words with two entries are geographical locations in Taiwan,  and we add extra word types to reflect the multiple entries in Wikidata.  For instance, 花蓮 \textit{huā lián} are found with 花蓮市 \textit{huā lián shì} (`Hualien City') and 花蓮縣 \textit{huā lián xiàn} (`Hualien County'). Therefore, they are divided into two word types corresponding to the two entries in Wikidata. After referencing Wikidata, the original 370-word-list has 38 words removed and extra 15 words added, resulting in a 347-word list.

After associating the words with Wikidata entries, we could identify the semantic category of each proper noun and determine its dot objects. We use Wikidata's ontological relations, {\it instance\_of} and {\it subclass\_of}, to find each word's semantic category and map it into corresponding dot objects. The relationships between Wikidata's category and dot objects are guided by the regular polysemy literature and annotators' lexical knowledge. Take the word 星巴克 {\it xīng bā kè} (`Starbucks') for example. In the Wikidata ontology, the semantic category for Starbucks is `business.' The company names are the examples of dot object {\texttt producer.product} in \citet{Pustejovsky2005}, where {\it Honda}, {\it IBM}, and {\it Microsoft} are among the instances of this dot object. Therefore, we could explicitly connect the Wikidata `business' category to the dot object {\texttt producer.product}. Another example is the dot object {\texttt organization.physical.information} in \citep{Pustejovsky2005,Vicente2021}, where the instances are {\it magazine}, {\it newspaper}, and {\it journal}. Similarly, we thus consider the `mass media' category to be the corresponding dot object. Once the connections are established, we trace all the words having the mapped categories in Wikidata and classify them accordingly. In this case, the words such as 特斯拉 {\it tè sī lā} (`Tesla'), 肯德基 {\it kěn dé jī} (`KFC'), 蘋果 {\it píng guǒ} (`Apple'), are all under the `business' category and would thus be categorized into the dot object \texttt{producer.product.location}. Table~\ref{tab:wikidata-c} shows the association between dot objects and the Wikidata category built in this step. 


\begin{table}[t]    
    \small
    \centering
    \caption{Connections between Wikidata Categories and Dot Objects.} \label{tab:wikidata-c}
    \begin{tabular}{lL{.5\textwidth}r}
    \toprule
    \textrm{Dot Object} & Wikidata Category & \makebox[1cm][r]{Num. of Words} \\
    \midrule
    Info*Phy & work of art & 4\\
    Loc*Org& human-geographic territorial entity, government agency, industrial zone, 
             intergovernmental organization, court. Chinese temple, museum, & 187\\
    Org*Hum & political party, military unit, sports organization, 
              religious organization, religious identity  & 39\\
    Org*Info*Phy*Hum & mass media & 22\\
    Org*Loc*Hum & university, educational institution, organization, hospital & 31\\
    Phy*Evt*Hum & award & 7\\
    Prcr*Prct*Loc & business & 57\\
    \bottomrule    
    \end{tabular}
\end{table}

\subsubsection{Annotating regular polysemy}
Considering the uneven distribution among the seven dot objects, we select 130 proper nouns to cover all the dot objects more evenly as our annotation targets. For every 130 words, we randomly sample 30 sentences and compile a dataset of 3900 sentences. For each sentence containing the target word, there will be a Wikidata category and a corresponding dot object, which then serve as the options for multiple-choice questions in the annotation process.

Four annotators are recruited to annotate the type class best represents the sense of the target word in a given sentence. For example, in the case of {\it xingbake} (`Starbucks'), the annotators are presented with three choices, namely  \texttt{producer}, \texttt{product}, and \texttt{location} in each of the 30 sentences. Then, in compliance with the behavior and denotation of the word `Starbucks,' annotators need to determine if this `Starbucks' under the specific context acts more like \texttt{producer}, \texttt{product}, or \texttt{location}. Some of the annotation samples are presented below. In (5), the target word {\it xīng bā kè} (`Starbucks') denotes an organization, or more specifically, a company with a CEO, and should be annotated as {\it producer}. In the context of (6), the target word {\it xīng bā kè} is `drinkable' since it is the direct object of the predicate 喝 {\it hē} (`drink'), and we can infer that it refers to the {\it product} sold by the company. Lastly, the target word {\it xīng bā kè} denotes a physical {\it location} in which people could `take a rest' in the sentence (7). In (8) and (9), we examine another dot object \texttt{organization.human}. In the former instance, the target word 海軍{\it hǎijūn }(`navy') plays the role of an intentional agent that is able to dispatch a frigate and gives orders. It's therefore being annotated as \texttt{organization}. On the other hand, the same target word in (9) does not stand for the group of military members or the organization as a whole but refers to certain individuals that are tested positive after long quarantine. Thus, they refer to the \texttt {human} dot object.


\begin{enumerate}
    
    \item 日本 星巴克 執行長 水口貴文 在 新聞稿 中 指出 \\ 
    rìběn xīngbākè zhíxíngcháng shuǐkǒuguìwén zài xīnwéngǎo zhōng zhǐchū\\
    Japan Starbucks CEO Takafumi.Minaguchi in newspaper inside point.out
    ‘The CEO of Starbucks in Japan, Takafumi Minaguchi, points out in the newspaper…’\vspace{4mm}
    
    \item 天天 喝 星巴克 總比 天天 吃 牛排 還好 吧\\
    tiāntiān hē xīngbākè zǒngbǐ tiāntiān chī niúpái háihǎo bā\\
    every day drink Starbucks better.than every day eat steak better AUX
    ‘And it's better to drink Starbucks every day than eat steak every day.’\vspace{4mm}
    
    \item 台北市區 某間 星巴克 稍做 休息\\
    táiběishìqū mǒujiān xīngbākè shāozuò xiūxí\\
    Taipei.city some Starbucks a.little rest
    ‘take a rest at a Starbucks in Taipei.’\vspace{4mm}
    
    \item 海軍 出動 成功艦 從 基隆 出海 待命\\
    hǎijūn chūdòng chénggōngjiàn cóng jīlóng chūhǎi dàimìng\\
    navy dispatch Cheng.Kung.Class.Frigate from Keelung launch.out stand.by
    ‘The navy dispatches Cheng Kung Class Frigates to stand by, launching out from Keelung.’\vspace{4mm}
    
    \item 海軍 之前 二採 陰性, 隔離 這麼 久 到 現在 還有 測出 陽性 的\\
    hǎijūn zhīqián èrcǎi yīnxìng gélí zhème jiǔ dào xiànzài háiyǒu cèchū yángxìng de\\
    navy before two.consecutive negative.tests quarantine such long until now still test positive de
    ‘Previously the navy have two consecutive negative tests and have been in quarantine for such a long period. Still, some of them are tested positive even until now.’\vspace{4mm}
\end{enumerate}

The four annotators are split into two groups, each responsible for two different datasets. As a result, there are 134 sentences considered ambiguous, incorrectly tokenized, or provided with insufficient context. Moreover, 65 sentences are annotated as {\it dot}, meaning that the target words display co-predication in the sentence. Although the phenomena of co-predication, simultaneous predictions that select two different senses in one single sentence, exhibit the very characteristics of dot objects in regular polysemy, it is not the major concern in the present study. Another concern is that most extracted sentences contain more than one target word. In this case, the annotators are asked to divide one sentence into multiple so that each instance focuses on the sense of the only target word. Thus, the final dataset consists of 4128 instances, eliminating the 134 sentences that could not be judged even by human raters and the 65 sentences that are considered `dot,' and expanding due to the frequently appeared instances with multiple target words.

The inter-rater agreements are calculated in two ways. On the one hand, pooling all the dot type classes together, the consistency of the annotation results for the two groups reach 83\% and 89.5\%. On the other hand, the rating agreement of dot objects is examined separately for a more careful calculation of annotation reliability. In addition, the inter-rater agreements in each dot object could also function as an indicator for further error analysis.

\begin{table}[t] 
    \centering
    \caption{Fleiss' $\kappa$ within Categories.}
    \label{tab:kappa}  
    \begin{tabular}{lrrr}
        \toprule
        Dot objects & Instances & Raters & $\kappa$ value\\
        \midrule
        Info.Phy & 93 & 2 &  0.726\\
        Loc.Org & 1250 & 2 & 0.647\\ 
        Org.Hum & 484 & 2 & 0.795\\
        Org.Info.Phy.Hum & 407 & 2 & 0.539\\
        Org.Loc.Hum & 685 & 2 & 0.639\\
        Phy.Evt.Hum & 205 & 2 & 0.679\\
        Prcr.Prct.Loc & 1202 & 2 & 0.769\\
        \bottomrule
    \end{tabular}    
\end{table}

We compute the Fleiss' $\kappa$, a statistical measure to assess agreement for multiple choice questions (Table \ref{tab:kappa}). It is found that the annotation results mostly fall between the interval of `substantial agreement,' that is, between .61 and .80. The only exception is the dot object \texttt{organization.information.physical.human}, being on the level of `moderate agreement' (between .41 and .60). However, it is also noticeable that this is the only class with four options included, indicating that the polysemous words in this category tend to be complicated.

\begin{table} 
    \small
    \centering
    \caption{Distribution of the Entity Type.} 
    \label{tab:annoed-dist}
    \begin{tabular}{rrrrrrrrr}
    \toprule
    Entity Type & Evt & Hum & Info & Loc & Org & Phy & Prcr & Prct\\
    \midrule
    GPE & 0 & 7 & 0 & 407 & 686 & 0 & 0 & 0\\
    LOC & 0 & 1 & 0 & 10 & 17 & 0 & 0 & 0\\
    ORG & 0 & 131 & 15 & 105 & 1176 & 1 & 646 & 176\\
    PRODUCT & 0 & 0 & 1 & 0 & 1 & 2 & 69 & 212\\
    WORK-OF-ART & 92 & 14 & 98 & 0 & 122 & 139 & 0 & 0\\
    \midrule
    Total & 92 & 153 & 114 & 522 & 2002 & 142 & 715 & 388\\
    \bottomrule    
    \end{tabular}
\end{table}

Table \ref{tab:annoed-dist} further tabulates the frequencies of each dot type class and named entity class. Two observations could be made from the table. First, the distributed frequencies in each row (named entity class) signify the lexical ambiguities of proper nouns, even after named entity recognition and entity linking. For example, when a proper noun is recognized as an \texttt{organization} (\texttt{Org} for short) entity, aside from denoting an organization, it could still refer to other dot type classes, such \texttt{producer} and \texttt{product}. Second, the distribution of dot type classes is skewed to \texttt{Org} class, partly resulting from its large number of named entities and the frequent usage of \texttt{Org} dot type class.

\section{Models and Experiments}
\subsection{Model architecture and inputs}


We formulate our WSD on common words and proper nouns as a multiple-choice QA task, following GlossBERT \citep{Huang2019} with weak supervision signals to train an integrated classifier.

\subsubsection{GlossBERT architecture and inputs} 
GlossBERT is first introduced to solve the WSD task, in which a BERT model is given a context $\{w_1, ..., w_m\}$ within which some target words $\{w_i1, ..., w_ik\}$ are specified. The model is then trained to find the most suitable entry in each $w_{ij}$'s predefined sense inventory. GlossBERT proposes to construct context-gloss pairs from the sense inventory of the target. These context-gloss pairs are constructed with the following steps. (1) For each target word within the sentence, extract the glosses of the $N$ possible senses of the target. (2) With the sentence denoted as \texttt{CONTEXT}, and the sense gloss denoted as \texttt{GLOSS}, the context-gloss pairs are constructed as the concatenation of {\texttt{[CLS] CONTEXT [SEP] GLOSS [SEP]}.}, where \texttt{[CLS]} and \texttt{[SEP]} are model-specific special tokens.

In other words, for each sentence, the number of context-gloss pairs produced is the same as the number of the target word's senses in the sense inventory. The model aims to predict the probability distribution for all context-gloss pairs belonging to the same context and choose the one with the highest probability. \citet{Huang2019} experimented with different schemes of constructing the pairs. They concluded that adding angular brackets, which act like a weakly supervised hint, around the target word in the test sentence boosts model performance the most among all settings.

Furthermore, to better guide the sense disambiguation task, we leverage the target word's part-of-speech (POS) tag. Specifically, we assume the target word's POS tag has been determined in the prior stage of the pipeline. We organized the POS tags of the ground-truth as well as the candidate senses from the 44 CKIP POS tags\footnote{https://ckip.iis.sinica.edu.tw/CKIP/paper/Sinica\%20Corpus\%20user\%20manual.pdf} to four categories: proper nouns (\textit{Nb}), common nouns (\textit{N}), verbs or verbal objects (\textit{V}) and others (\textit{OTHERS}). Then, we kept the candidate senses with the same simplified POS tags of the target words. This setting is called \texttt{POS-guided} setting. Table \ref{tab:dataset-stat}'s first rows display the number of examples and sequences of the WSD dataset, in which example number denotes the number of example sentences; sequence number denotes the number of all the candidate-sense sequences flattened out from all examples.

\subsubsection{Model inputs and architecture}
We construct our model input sequences as a set of context-gloss pairs. Let $T$ denotes the set of test sentences in the WSD dataset. For each test sentence $t_i \in T$, denoted \texttt{TEST-SENT}, one and exactly one target word $w_i$ (\texttt{TGT}) is identified, and its predefined sense inventory $sense\_inv_i$ is retrieved using \verb|Chinese Wordnet 2.0|~\footnotemark. Next, for each sense $s_{ij} \in sense\_inv_i$, we retrieve its definition (\texttt{SENSE-DEF}) and randomly select one example sentence of the sense (\texttt{SENSE-EX-SENT}). For a context-gloss pairing common word WSD, the input sequence is concatenated as:
{\texttt{[CLS] TEST-SENT [SEP] TGT, SENSE-DEF, SENSE-EX-SENT [SEP]}.} Different from the original GlossBERT, we add more information in the \texttt{GLOSS} part to allow the model to leverage more information. Each $t_i$ gets the same number of context-gloss pairs as the number of its candidate senses. Only one of them is labeled as the correct answer, i.e., True. Those $t_i$ whose $w_i$ has less than two predefined senses are discarded. 

\footnotetext{CWN 2.0 is an extension of CWN in both sense data coverage, gloss re-articulation and others. Data is browsable and available at \url{https://lopentu.github.io/CwnWeb/}}

\noindent 
On the other hand, the WSD for proper nouns of regular polysemy (RP) follows a similar procedure to construct context-gloss pairs. $T'$ denotes the set of test sentences of RP. For each test sentence $t_i \in T'$ (\texttt{TEST-SENT}), one target word $w_i$ is identified and its predefined RP-class inventory $rpclass\_inv_i$ is retrieved. Next, for each type class $c_{ij} \in rpclass\_inv_i$ (\texttt{RPCLASS}), we compose its corresponding gloss (\texttt{RPCLASS-GLOSS}) according to \textit{Revised Mandarin Chinese Dictionary by Ministry of Education, R.O.C} \footnote{Data are accessible via \url{https://dict.revised.moe.edu.tw}} (See Table \ref{tab:rpgloss} for the RP classes and corresponding glosses). For each test sentence $t_i$, it has a $|rpclass\_inv_i|$ number of context-gloss pairs. The RP input sequence is concatenated as: {\texttt{[CLS] TEST-SENT [SEP] TGT, RPCLASS, RPCLASS-GLOSS [SEP]}.} Similar to WSD inputs, only one context-gloss pair is labeled as the correct answer. Finally, all target words in the sequences are marked with angular brackets as weak supervision signals. Table \ref{tab:inputseq} presents the context-gloss pairs corresponding to one test sentence for WSD and RP, respectively. \footnote{In each context-gloss pair, the target word is marked with weak supervision bracket ($\langle$ $\rangle$). The \textit{context} is the sentence containing the target word to be disambiguated; the \textit{gloss} comprises the gloss and an example sentence in CWN (in WSD), or the gloss of the type classes (in RP disambiguation). The context in the WSD pair is loosely translated as ``The employer made a $\langle$ complaint $\rangle$ to the superior...'', and the four candidate senses are (1) from speaker's perspective; (2) the exterior shape; (3) a lawsuit; (4) complaint. The context in the RP pair is translated as ``For the tuition of Harvard, he...'', the four candidates of the regular polysemy are organization, location, and human.}

\begin{table}[t]
    \small
    \begin{center}
    \caption{RP class and corresponding gloss}
    \label{tab:rpgloss}
    \begin{tabular}{llll}
    
        \toprule
        RP Type-Class & \parbox[c]{1cm}{\centering Chinese \\ Translation} &
            Chinese Gloss\\
        \midrule
        Physical & 有形的 & \parbox[c]{6cm}{\vspace{4pt}
                            有具體形狀。\\
                            \emph{tangible objects.}
                            \vspace{4pt}}\\
        Organization & 機構 & \parbox[c]{6cm}{\vspace{4pt}
                                泛指機關團體或工作單位。\\
                                \emph{general reference to administrative and functional structures.}
                                \vspace{4pt}}\\
        Producer &作者;製造商 & \parbox[c]{6cm}{\vspace{4pt}
                                創作詩歌、文章或其他藝術品的人;\\
                                製造或出售各種物品的商家。\\
                                \emph{creators of poetry, articles, or other artworks; business who produces or supplies goods or services.}
                                \vspace{4pt}}\\
        Product & 作品;產品 & \parbox[c]{6cm}{\vspace{4pt}
                            文學藝術方面創作的成品;\\
                            生產的物品。\\
                            \emph{artistic creations; commodities that have been produced.}
                            \vspace{4pt}}\\
        Location & 地點 & \parbox[c]{6cm}{\vspace{4pt}
                            所在的地方。\\
                            \emph{positions or occupied sites.}
                            \vspace{4pt}}\\
        Event & 事件 & \parbox[c]{6cm}{\vspace{4pt}
                        事情、事項。\\
                        \emph{circumstances, incidents.}
                        \vspace{4pt}}\\
        Human & 人類 & \parbox[c]{6cm}{\vspace{4pt}
                        人的總稱。\\
                        \emph{general term of humanity.}
                        \vspace{4pt}}\\
        Information & 資訊 & \parbox[c]{6cm}{\vspace{4pt}
                            泛指一般資料和訊息。\\
                            \emph{general reference to data, knowledge, and messages.}
                            \vspace{4pt}}\\
        \bottomrule
        
    \end{tabular}
    \end{center}
\end{table}

\begin{table}
    \begin{center}
    \caption{The format of the context-gloss pairs in WSD and regular polysemy disambiguation task.}
    \label{tab:inputseq}
    \small
    \begin{tabular}{p{11.5cm}l}
        \toprule
        WSD context-gloss pairs & Label \\
        \midrule 
         \pcr{[CLS]}雇主一〈狀〉告到上頭 ... \pcr{[SEP]} \umark{狀}:以說話者的觀點，...,...。\pcr{[SEP]}&False\\
         \pcr{[CLS]}雇主一〈狀〉告到上頭 ... \pcr{[SEP]} \umark{狀}:構成特定對象的外部輪廓。, ... 。\pcr{[SEP]}& False\\ 
         \pcr{[CLS]}雇主一〈狀〉告到上頭 ...  \pcr{[SEP]} \umark{狀}:請求司法機關審理的案件。, ...。 \pcr{[SEP]}& False\\
         \pcr{[CLS]}雇主一〈狀〉告到上頭 ... \pcr{[SEP]} \umark{狀}:刻意向較有權力的人..., ...。\pcr{[SEP]}& True\\
        \bottomrule
    \end{tabular}
    \vspace{\baselineskip}
    \small
    \begin{tabular}{p{11.5cm}l}
        RP context-gloss pairs & Label \\
      \midrule 
      \pcr{[CLS]}他最近為了〈哈佛〉學費...\pcr{[SEP]}\umark{哈佛}:機構,泛指機關團體...。\pcr{[SEP]}&True\\
       \pcr{[CLS]}他最近為了〈哈佛〉學費...\pcr{[SEP]}\umark{哈佛}:地點,所在的地方。\pcr{[SEP]}&False\\
       \pcr{[CLS]}他最近為了〈哈佛〉學費...\pcr{[SEP]}\umark{哈佛}:人類,人的總稱。\pcr{[SEP]}&False\\
      \bottomrule
    \end{tabular}
    \end{center}
\end{table}

Our model is based on the pretrained \textit{bert-base-chinese}, which has a fully-connected layer for the classification task. 
We split the WSD dataset into \textit{WSD\_train, WSD\_test}; RP dataset into \textit{RP\_train, RP\_test}. In the experiment setting, \textit{WSD\_train} and \textit{RP\_train} are pooled and trained together in training phase. When fine-tuning, we flatten each example (i.e., a set of context-gloss pairs belonging to the same test sentence) into context-gloss pair (sequences), and consider each context-gloss pair as one sequence (see Table \ref{tab:dataset-stat}). In this way, the batch size denotes \textit{the number of context-gloss pairs} sent into the model at one time rather than the number of multiple-choice examples. The batch size is set to 16; the initial learning rate $1e-5$, and the model is trained for two epochs. We use AdamW\citep{Loshchilov2017} optimizer and a linear scheduler with a warmup at the first 100 steps.


\begin{table}[ht]
    \centering
    \caption{Dataset Statistics}
    \label{tab:dataset-stat}        
    \begin{tabular}{llrr}
        \toprule
        \multicolumn{2}{l}{Dataset} & Train Split  & Test Split \\
        \midrule[0.4pt]
        \multirow{2}{*}{WSD} & Examples  &  36,622 & 9,162 \\
                          & Sequences &  386,351 & 83,418 \\
        \midrule[0.2pt]
        \multirow{2}{*}{RP} & Examples    & 3,641      & 866      \\
                            & Sequences   & 9,227   &    2,333    \\

        \bottomrule
    \end{tabular}    
\end{table}

\subsection{Results and discussions}
\begin{figure}
\centering
\begin{minipage}{.5\textwidth}
  \centering
  \includegraphics[width=.93\linewidth]{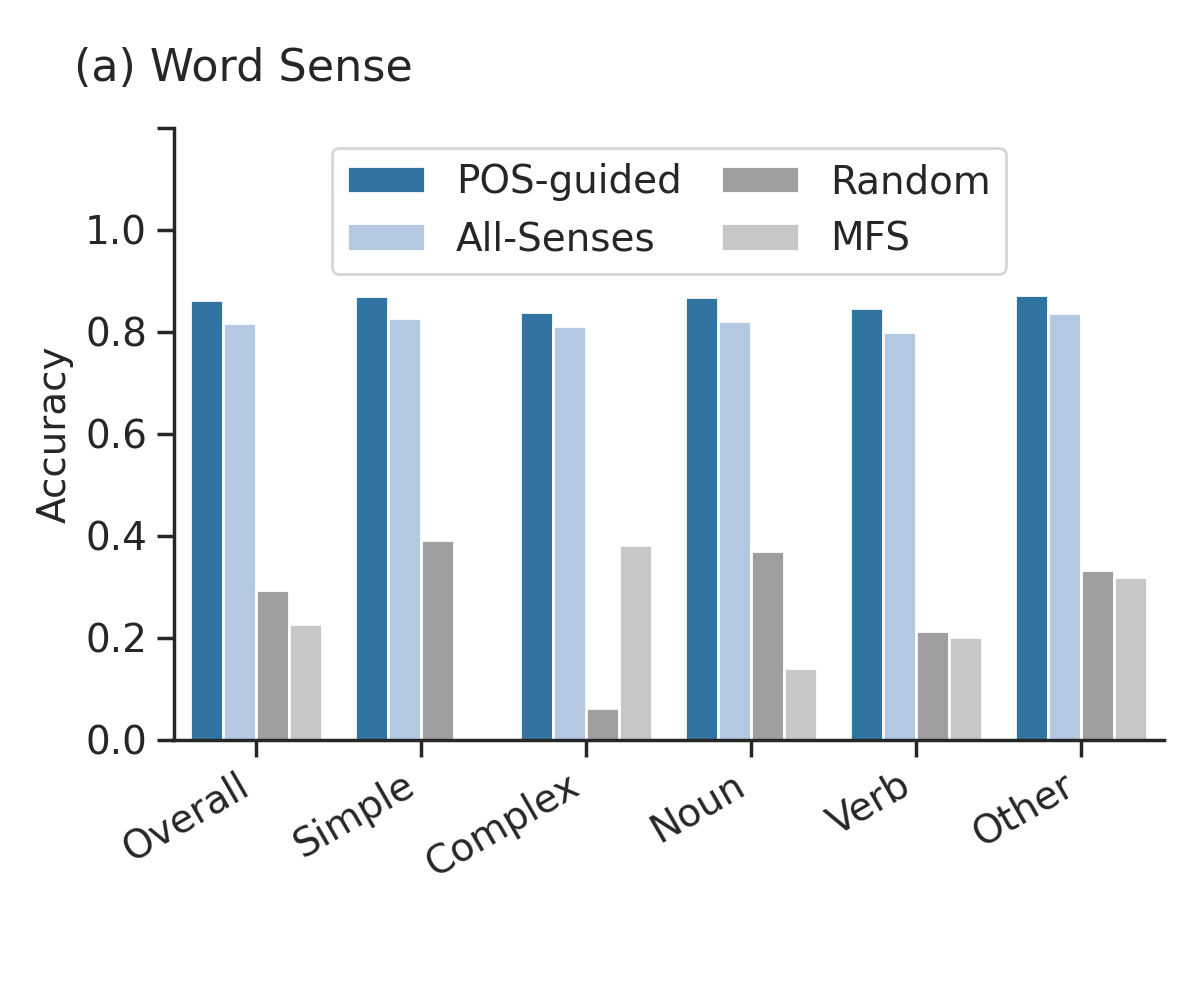}
  
\end{minipage}%
\begin{minipage}{.5\textwidth}
  \centering
  \includegraphics[width=.93\linewidth]{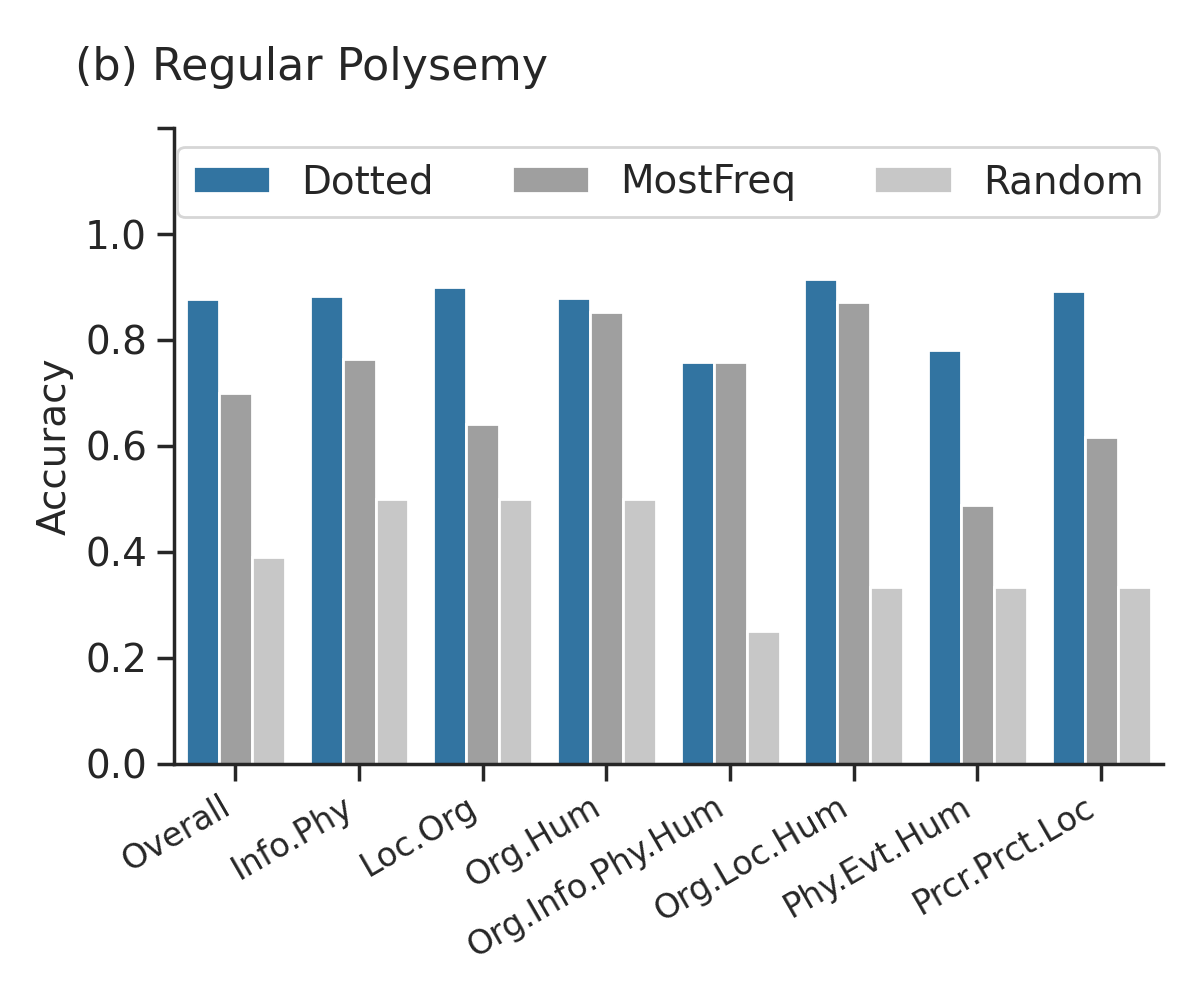}
\end{minipage}
\vspace*{5mm}
\caption{Evaluations of word sense disambiguation and regular polysemy disambiguation. The POS-guided, All-senses, Dotted denote our inference settings.}
\label{fig:evals-wsd}
\end{figure}

The model's prediction accuracies under different conditions are shown in Figure \ref{fig:evals-wsd}. Overall, the model achieves .86 accuracy in the WSD task and .88 in the RP task respectively. By contrast, the random baseline of the WSD task is .29, and the one of the RP task is .39.

In addition to the overall model performance, we further compare the accuracy with two other baselines: All-senses and Random. In the All-senses baseline, the model needs to predict the target sense of a word from all the word's senses in the sense inventory. This baseline measures how the current model is dependent on the POS hints. On the other hand, the random baseline selects from the word's sense inventory by chance. The results (Figure \ref{fig:evals-wsd} a. ) show the overall accuracy of All-sense baseline is .82, and the random baseline's accuracy is .29. While the model does benefit from the POS hints (by 4\%), the small difference in accuracy indicates that it actually learns how to disambiguate common words and does not entirely depend on POS filter. 

We also compare model performances under different types of words. The aims are to illustrate how the model is affected by factors such as the number of possible candidates or lexical categories (e.g., the word's POS). First, we compare the accuracies under the different numbers of possible candidates. In particular, we divide the WSD test dataset into two subsets: 3,691 simple words, words having ten or fewer senses, and 5,461 complex words, words having more than ten senses. The average number of possible candidates for the simple words is 4.52, and 15.48 for the complex ones. The difference could also be observed from the chance level accuracies, i.e., 39\% for simple words and 6\% for complex ones. Nevertheless, the model achieves comparable accuracy in simple and complex words, regardless of POS-guided (87\% for simple words and 84\% for complex ones) or All-senses (83\% for simple and 81\% for complex) conditions. Secondly, to investigate the possible influences of lexical categories, we also divide the WSD test set into three subsets by the word's POS, namely, 2,289 nouns, 3,639 verbs, and 3,224 other words. The average number of sense candidates is 10.98 for nouns, 17.78 for verbs, and 16.09 for other words. However, the accuracies of the three word types are similar. While the noun scores higher than the verb by 2\%, it might result from the number of sense candidates rather than the lexical factors of verbs. These results show that our model disambiguates word's senses guided by POS information but not entirely dependent on POS hints.

The characteristic of this dataset is evident when considering the most frequent-sense (MFS) accuracies. The MFS accuracies in the literature are usually hard to beat as the dataset comes from the sense-annotated corpus. The most frequently used sense usually accounts for a large portion of the words used. However, most of our dataset comes from the example sentences; only those from the additional annotation data would have a more skewed sense distribution. As we only include complex words for annotation, the MFS accuracy is, therefore, higher for the complex condition (38\%), which is consistent with the MFS patterns in the literature. It is also interesting to note that the MFS accuracy of the Simple condition is 0\%. It results from no annotation data being present in the Simple condition; thus, all senses are equally likely. Furthermore, the CWN extracted data has only one instance, either for training or testing for each sense, and one could not use the MFS strategy to predict any sense of simple words correctly. Overall, the MFS patterns again demonstrate that the model is robustly learned to predict intended word uses.

\begin{table} 
    \small
    \centering
    \caption{Evaluation Statistics of word sense disambiguation. The underline indicates the inference setting of our WSD model.} 
    \label{stats-WSD}
    \begin{tabular}{rrrrrrr}
    \toprule
    Model & Overall & Simple & Complex & Noun & Verb &Other\\
    \midrule
    \underline{{\bf POS-guided}} & .86 & .87 & .84 & .87 & .85 & .87 \\
    \underline{All-Senses} & .82 & .83 & .81 & .82 & .80 & .84 \\
    Random & .29 & .40 & .06 & .37 & .21 & .33 \\
    MFS & .23 & .00 & .38 & .14 & .20 & .32 \\
    \bottomrule    
    \end{tabular}
\end{table}

\begin{table}
\small
    \centering
    \caption{Evaluation statistics of regular polysemy. The underline indicates the inference setting of our WSD model. Type classes are abbreviated with their initials: I: Information, Ph: Physical, L: Location, H: Human, O: Organization, E: Event, Pr: Producer, Pt: Product. } 
    
    \label{stats-RP}
    \begin{tabulary}{1.01\linewidth}{p{1.15cm}CCCCCCCC}
        \toprule
        Model & Overall & I*P & L*O & O*H & O*I*Ph*H & O*L*H & Ph*E*H & Pr*Pt*L\\
        \midrule
        \underline{{\bf Dotted}} & .88 & .88 & .90 & .88 & .76 & .91 & .78 & .89\\
        MostFreq & .70 & .76 & .64 & .85 & .76 & .87 & .49 & .62\\
        Random & .40 & .50 & .50 & .50 & .25 & .33 & .33 & .33\\
        \bottomrule
    \end{tabulary}
\end{table}

Similarly, we also examine the model performances on regular polysemy. In addition to our model, we compute the accuracies of a most-frequent baseline and a random baseline. The goal is to show the model learn to differentiate regular polysemy in addition to the class priors. Specifically, we examine the accuracies of each dot-type category in our model and the baselines. The results are shown in Figure \ref{fig:evals-wsd}b. The overall model accuracy is 17\% higher than the most-frequent baseline (70\%) and in nearly all dot-type categories. The only exception is the \texttt{Org.Info.Phy.Hum} category, where the model accuracy is the same with always predicting the most-frequent one (i.e., \texttt{Org}). In a closer examination, the category is the most complicated 4-class-dot-type, and the class distribution is highly skewed. Among the 83 instances, there are 63 \texttt{Org}s, 10 \texttt{Info}s, 6 \texttt{Hum}s, and 4 \texttt{Phy}s. The same patterns are also observed in other highly skewed categories, such as \texttt{Org.Hum}, \texttt{Org.Loc.Hum}. However, for other more evenly-distributed categories, the model achieves higher accuracy than the most-frequent baselines. 

\begin{figure}
\centering
  \centering
  \includegraphics[width=.5\linewidth]{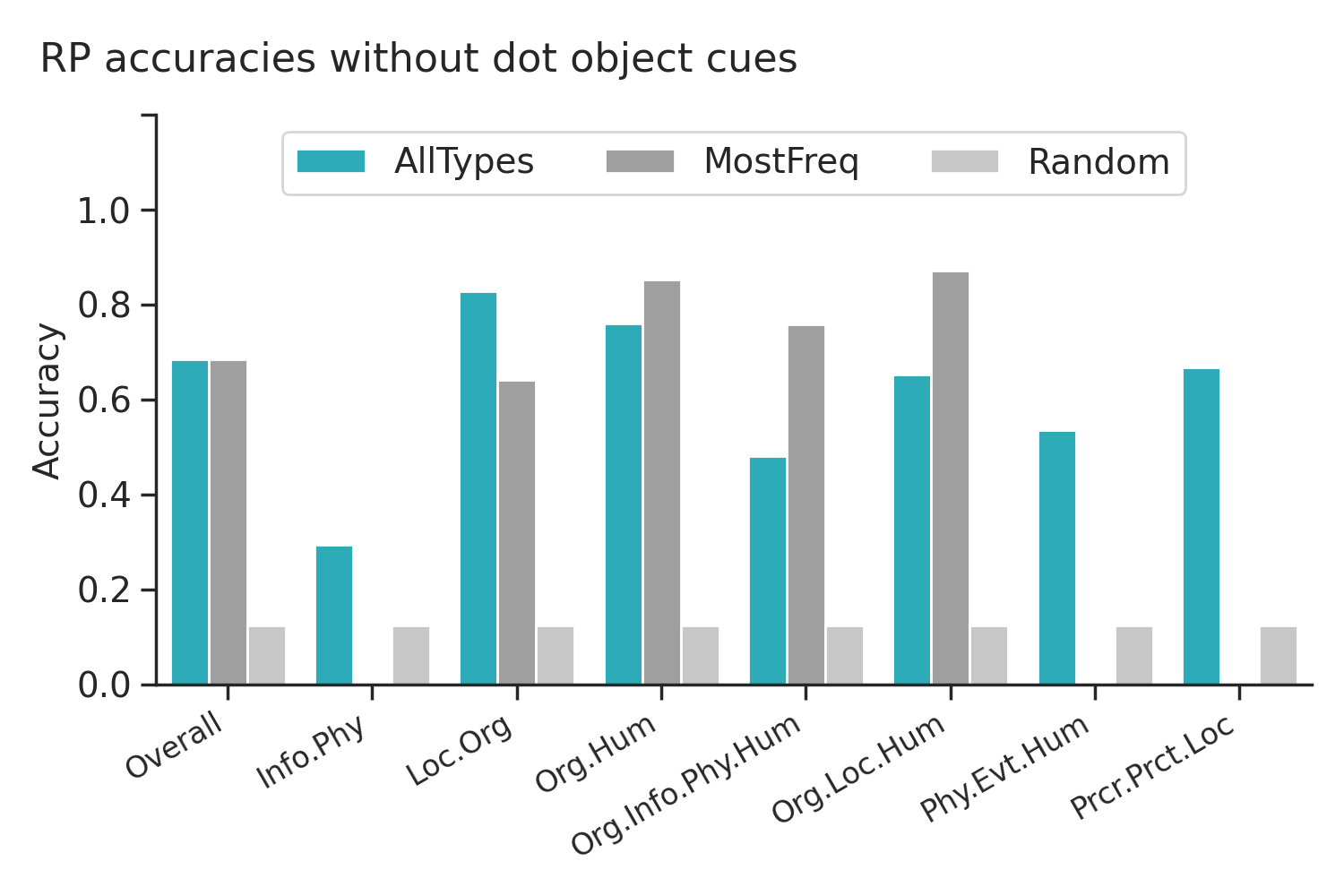}
\caption{Regular polysemy disambiguation without dot object cues.}
\label{fig:rp-all-types}
\end{figure}

Furthermore, we also evaluate how the model depends on the dot object information. Figure \ref{fig:rp-all-types} presents the accuracies of proper noun disambiguation without the constraints of dot objects. That is, a \texttt{Info.Phy} dot object is no longer a binary classification, but a multi-class one (8 dot-type classes, \texttt{AllTypes} condition). The most-frequent class (\texttt{MostFreq}) and \texttt{Random} accuracies are provided as baselines. While due to the extremely skewed distribution of dot-type classes, the categories involving the \texttt{Org} class have very high \texttt{MostFreq} accuracies. Nevertheless, other categories show performance significantly better than the random baseline. These results indicate that dot object cues are important to regular polysemy disambiguation. However, aside from the skewed distribution, the model does capture the regular polysemy in the running texts.

\section{Conclusion and future works}

This paper presents a combined WSD model which is capable of polysemy resolution for both common words and proper names. Due to the co-existing property of referential multiplicity and lexical diversity, the named entities represent a unique NLP object that calls for a disambiguation strategy from a linguistic perspective.

Our proposed model features in that, besides relying on an enumerated sense inventory for the common word WSD, it addresses the aspect of lexical ambiguities of proper names from the perspective of regular polysemy. While our gloss-based model follows the previous model architecture \citep{Huang2019}, the architecture enables us to flexibly leverage the glosses and example sentences in CWN and provides a unified way for training and inferences. We show that the model identifies the intended uses of common and proper words, even from sparse training data. The overall accuracies are 86\% for common words and 88\% for proper names. The model not only serves as a practical NLP tool but plays an essential role in lexical resource development, such as helping make sense distinctions on new words and extend semantic relations. Specifically, a good WSD model is crucial to connect the contextualized embeddings to the sense inventory, which is helpful for a better WSD model and a complete lexical resource.

One of the future works is to improve how proper names are assigned to different dot objects. We have already seen that Wikipedia is a valuable resource in WSD and serves as a reference ontology in this study. Thus, we could leverage the semantic class in Wikipedia to assign, either through algorithmic rules or classification models, dot objects with less human interventions. 

Other future works include extending regular polysemy to common words. In this paper, we use dot objects as notations and theoretical tools to operationalize regular polysemy of proper names; in fact, common words could be regularly polysemous, as revealed in CoreLex \citep{Buitelaar1998} analysis, where the senses in PWN and 126 semantic types of systematic polysemy could be derived. The patterns may also be observed in Chinese lexical semantics and leverage the systematic relations to provide another perspective of sense granularity.


\end{CJK*}
\end{document}